\pdfoutput=1

\documentclass[11pt]{article}

\usepackage[]{acl}

\usepackage{times}
\usepackage{latexsym}
\usepackage{array}
\usepackage{authblk}

\usepackage[T1]{fontenc}

\usepackage[utf8]{inputenc}

\usepackage{microtype}
\usepackage{hyperref}

\usepackage{inconsolata}
\usepackage{amsthm}
\newtheorem{definition}{Definition}
\usepackage{listings}
\usepackage{fancybox}
\usepackage{graphicx}
\usepackage{caption}
\usepackage{subcaption}
\usepackage{booktabs} 

\title{Lost in Transcription: Identifying and Quantifying the Accuracy Biases of \\ Automatic Speech Recognition Systems Against Disfluent Speech}

 \author{{\bf Dena Mujtaba\textsuperscript{1}, Nihar R. Mahapatra\textsuperscript{1}, Megan Arney\textsuperscript{1}, J. Scott Yaruss\textsuperscript{1}}, \\{\bf Hope Gerlach-Houck\textsuperscript{2}, Caryn Herring\textsuperscript{3}, and Jia Bin\textsuperscript{1} } \\
\textsuperscript{1}Michigan State University, \textsuperscript{2}Western Michigan University, \\ \textsuperscript{3}FRIENDS: The National Association of Young People Who Stutter \\

\textsuperscript{1}\texttt{\{mujtabad,nrm,arneymeg,jsy,binjia\}@msu.edu} \\
\textsuperscript{2}\texttt{hope.gerlach@wmich.edu} \\
\textsuperscript{3}\texttt{caryn@friendswhostutter.org}
 }

\begin{document}
\maketitle
\begin{abstract}
Automatic speech recognition (ASR) systems, increasingly prevalent in education, healthcare, employment, and mobile technology, face significant challenges in inclusivity, particularly for the 80 million-strong global community of people who stutter. These systems often fail to accurately interpret speech patterns deviating from typical fluency, leading to critical usability issues and misinterpretations. This study evaluates six leading ASRs, analyzing their performance on both a real-world dataset of speech samples from individuals who stutter and a synthetic dataset derived from the widely-used LibriSpeech benchmark. The synthetic dataset, uniquely designed to incorporate various stuttering events, enables an in-depth analysis of each ASR's handling of disfluent speech. Our comprehensive assessment includes metrics such as word error rate (WER), character error rate (CER), and semantic accuracy of the transcripts. The results reveal a consistent and statistically significant accuracy bias across all ASRs against disfluent speech, manifesting in significant syntactical and semantic inaccuracies in transcriptions. These findings highlight a critical gap in current ASR technologies, underscoring the need for effective bias mitigation strategies. Addressing this bias is imperative not only to improve the technology's usability for people who stutter but also to ensure their equitable and inclusive participation in the rapidly evolving digital landscape.
\end{abstract}

\section{Introduction}\label{S-intro}
\begin{figure}[t!]
\centering
\includegraphics[width=.99\columnwidth]{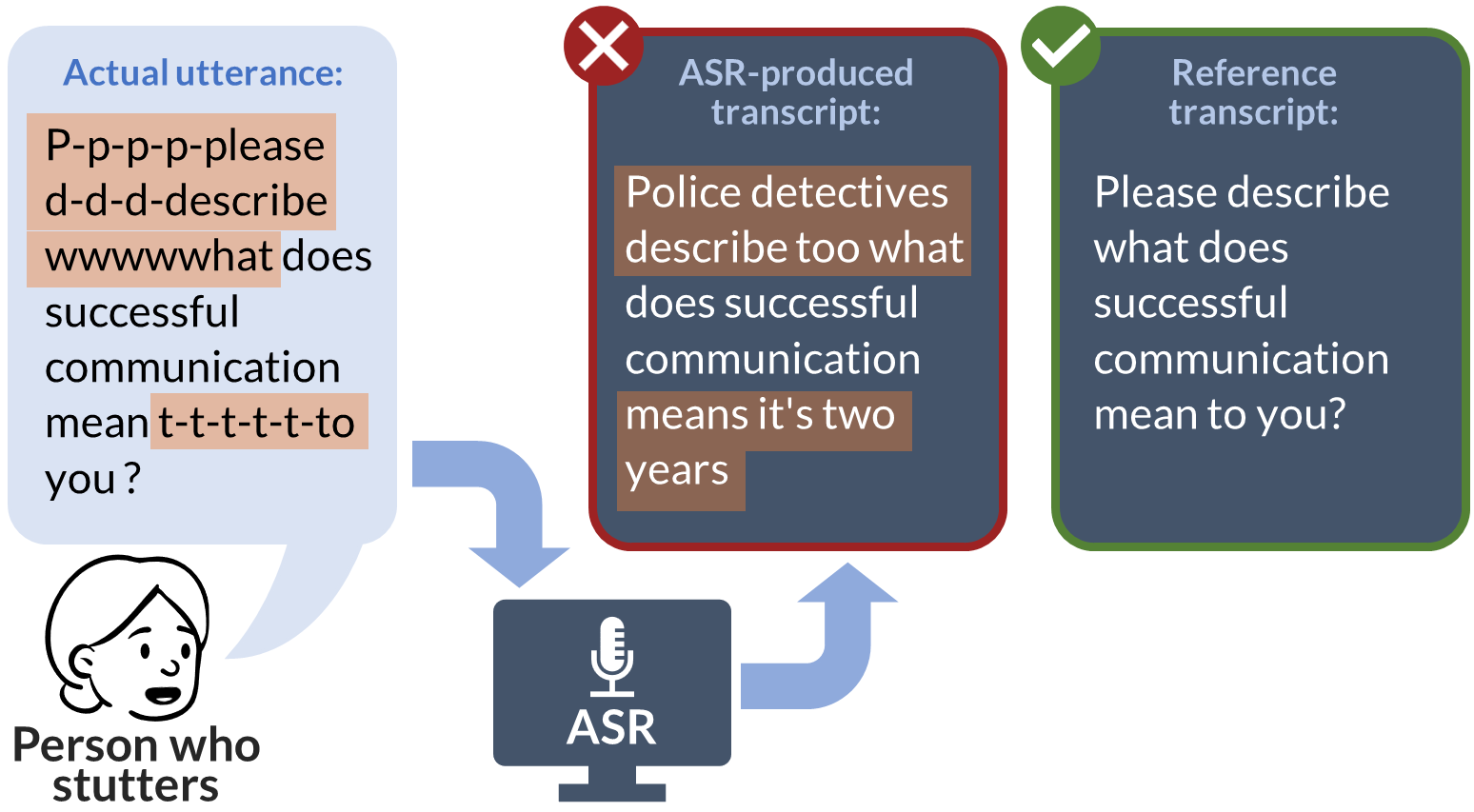}
\caption{Transcript produced by an automatic speech recognition (ASR) model from a speech sample of an individual who stutters, sourced from FluencyBank \cite{ratner2018fluency}. This figure illustrates a real-world scenario where the individual is conducting an interview. Highlighted are both the disfluencies characteristic of stuttered speech and the corresponding transcription inaccuracies produced by the ASR.}
\label{fig:example}
\vspace*{-.1in}
\end{figure}

\subsection{Background and Motivation}
Stuttering is a complex neurodevelopmental condition affecting over 80 million people worldwide \cite{wu2023world, yairi2013epidemiology}. This condition is characterized by various involuntary speech disruptions or disfluencies
\cite{johnson1959onset, ambrose1999normative}, including repetitions of parts of words (``li-li-like this''), prolongations of sounds (``lllllike this''), and blocks (silent periods, ``l---ike this''). While everyone may experience disfluencies, in people who stutter, these are more pronounced and are accompanied by a loss of control over speech \cite{tichenor2019stuttering}. Society's negative perceptions of stuttered speech often leads to discrimination and lower job satisfaction, as fluent speakers tend to be preferred in employment settings \cite{gerlach2018stuttering,shrm_discrimination,plexico2019influence}.

In the post-COVID-19 era, characterized by the widespread adoption of voice-activated artificial intelligence (\textit{voice AI}), individuals who stutter are confronted with heightened challenges, especially in employment contexts. Automatic speech recognition (ASR) systems, the fundamental technology underpinning voice AI and designed to transcribe spoken language into text \cite{prabhavalkar2023end}, have emerged as notable barriers in various aspects of daily and professional life for this demographic. The broad integration of ASR technology across sectors such as education, healthcare, and employment, as well as in mobile devices and other applications \cite{prabhavalkar2023end,malik2021automatic}, has led to the inadvertent marginalization of people who stutter. These systems frequently struggle to accurately decode speech that differs from typical patterns \cite{techexplore,wu2023world,lea2023user}, a fact exemplified in Fig.~\ref{fig:example}. In stark contrast, ASRs demonstrate up to 95\% word accuracy when processing standard speech \cite{tobin2022personalized}. The ramifications of this disparity are profound, ranging from difficulties in executing simple tasks, such as using automated phone systems for prescription refills, to experiencing bias in AI-driven job interviews \cite{junior2020person,center_2019,meyer2018amazon}.

As the global voice recognition market is projected to nearly quintuple by 2030, reaching a valuation of US\$59.62 billion \cite{fortune-bi}, the prevalence of AI-powered voice technologies, bolstered by the recent success of large language models, is set to increase dramatically \cite{horvitz2016one}. This growth suggests an impending escalation in bias against individuals who stutter. Leading technology companies like Apple and Google have begun addressing variations in speech, introducing alternative solutions such as typing commands to voice assistants \cite{techexplore}. However, these measures, while helpful, fall short of addressing the inherent bias in ASR accuracy. In particular, initiatives like Google’s Project Euphonia focus on collecting data for differing speech patterns, aiming to enhance ASR capabilities for understanding people with speech impairments like dysarthria \cite{tobin2022personalized}. However, these efforts often neglect the unique aspects of stuttered speech. Given the significant variability of stuttering across different individuals and situations \cite{tichenor2021variability}, this oversight reveals a critical gap. Unless ASRs are specifically tailored and refined to accommodate the diverse expressions of stuttering, they are likely to continue underperforming for individuals who stutter, exacerbating existing societal biases.

\subsection{Our Contributions}
This paper examines the response of contemporary ASRs to disfluent speech characteristic of stuttering, with a focus on identifying and quantifying biases. While previous research has explored speech variations such as dysarthria \cite{shor2019personalizing,moore2018whistle} and ASR fairness in the context of gender, age, or accent differences \cite{feng2021quantifying,liu2022towards}, biases specific to disfluent speech have remained largely unexplored. The work of Lea et al. \cite{lea2023user} investigates interactions between people who stutter and AI-based voice assistants, yet it stops short of providing a quantitative analysis of biases.

Overall, the main contributions of this paper relative to previous work are threefold: (1) Quantitative analysis of bias: We conduct a comprehensive analysis of bias in both the word error rate (WER) and character error rate (CER) of six popular ASRs, utilizing real speech data from people who stutter. This approach offers a deeper understanding of how these systems perform with speech patterns that differ from the norm. (2) Semantic accuracy evaluation: We extend our analysis to the semantic accuracy of these ASRs, exploring how well they preserve the meaning of disfluent speech, which is crucial for effective communication. (3) Synthetic disfluent speech dataset: To address the scarcity of stuttered speech datasets, we have developed a synthetic disfluent speech dataset using novel text-to-speech and audio manipulation techniques. This synthetic dataset, designed to complement real stuttered speech samples, allows us to finely control the types and frequencies of disfluencies in speech samples. This controlled environment enables a granular examination of how various disfluencies and the inherent variability in stuttering affect ASR performance. Our findings aim to inform bias mitigation strategies in ASR development, ultimately aiding people who stutter in their daily interactions with voice-driven technologies.

The remainder of the paper is organized as follows. In Section \ref{S-relatedwork}, we provide a literature review of state-of-the-art ASRs, ASRs for people with speech variations, and define fairness/bias measurement in the context of AI. Section~\ref{S-methods} details our methodology, datasets, models, and metrics for studying bias. Section~\ref{S-results} presents our findings and their analysis. Finally, Section~\ref{S-conclusion} concludes with remarks and suggestions for future research directions.

\section{Related Work}\label{S-relatedwork}
This section provides a brief literature review of three key areas to set the stage for subsequent sections examining bias in ASRs: the evolution and current state of ASR models, their adaptation for people with speech differences, and the measurement of bias in AI models.

\subsection{Automatic Speech Recognition}
Recent advancements in ASR have been largely driven by the integration of deep neural networks, leading to a significant reduction in WER compared to classical statistical multi-stage models \cite{prabhavalkar2023end}. These neural network models adopt an end-to-end (E2E) approach, directly mapping input acoustic features to their corresponding transcripts \cite{prabhavalkar2023end}. Key datasets like LibriSpeech \cite{panayotov2015librispeech}, encompassing 1,000 hours of diverse audiobook recordings, and TIMIT \cite{timi}, featuring recordings of 630 speakers across American English dialects, have been pivotal in training and benchmarking ASRs. Among the notable ASR architectures are the Conformer \cite{gulati2020conformer}, wav2vec 2.0 \cite{baevski2020wav2vec}, and Whisper \cite{radford2023robust}, with Whisper demonstrating remarkable zero-shot performance and the best WER at the time of this writing \cite{gulati2020conformer, baevski2020wav2vec, radford2023robust}. However, these models' effectiveness on stuttered speech remains underexplored. Comprehensive surveys on ASR systems are provided by Malik et al. \cite{malik2021automatic} and on E2E ASRs by Prabhavalkar et al. \cite{prabhavalkar2023end}.

\subsection{Automatic Speech Recognition for People with Speech Differences}
Efforts to enhance ASR performance for people with different speech patterns, such as stuttering, have primarily focused on disfluency identification. This involves classifying disfluency events to enhance ASR accuracy, either by adjusting, ignoring, or removing these elements before they are processed by an ASR, or by providing their location to the ASR \cite{zayats2016disfluency, shonibare2022enhancing}. Notable datasets supporting  stuttering identification include SEP-28K, KSoF, and LibriStutter, although they lack properly coded transcripts for ASR training and evaluation \cite{sheikh2022machine, lea2021sep, kourkounakis2020fluentnet}. While there is limited research on directly enhancing ASRs for stuttering, some progress has been made. Lea et al. fine-tuned Apple's Speech framework models with data from people who stutter, observing a reduction in WER. They also applied disfluency refinement and endpoint truncation techniques to improve accuracy \cite{lea2023user}. Mitra et al.'s work on tuning decoding parameters of ASRs also contributes to this effort \cite{mitra2021analysis}. Additionally, other research, such as those by Tobin et al., Shor et al., and MacDonald et al., has focused on adapting ASRs for various speech differences \cite{tobin2022personalized, shor2019personalizing, macdonald2021disordered,kim2008dysarthric, moore2020uncommonvoice}.

\subsection{Fairness in Artificial Intelligence}
Fairness in AI involves mitigating human and systemic biases that may arise from factors like unrepresentative training data, inappropriate features used during training, or biased target labels \cite{mujtaba2019ethical}. The concept of fairness in AI has evolved, with historical foundations in anti-discrimination legislation and contemporary frameworks like the AI Risk Management Framework by NIST aiming to enhance AI trustworthiness \cite{eeo, nist}. A fundamental fairness criterion in AI literature is ``fairness through unawareness''  \cite{kusner2017counterfactual}:

\begin{definition}
\textbf{Fairness Through Unawareness:}
\textit{An AI model $C$ is deemed fair under the principle of 'Fairness Through Unawareness' if it does not explicitly use protected attributes $A$ in its decision-making process. 
This principle is operationalized through a mapping $\hat{Y}{:}\; X \rightarrow Y$, where $X$ represents the set of observable features excluding any protected attributes $A$, $\hat{Y}$ represents a model's prediction, and $Y$ the intended outcome.}
\end{definition}

In this paper, $A$ represents the presence of disfluencies, $C$ is the ASR model under evaluation, $X$ is the user's speech input, and $Y$ is the transcript produced by an ASR. Our study investigates the correlation between the presence of disfluencies ($A$) and model error rates to evaluate ASR fairness. This can be captured through the notion of ``counterfactual fairness'' \cite{kusner2017counterfactual,wachter2017counterfactual}:

\begin{definition}
\textbf{Counterfactual Fairness:}
\textit{An AI model $C$ is deemed counterfactually fair if its predictions $\hat{Y}$ in the real world are the same as those in the counterfactual world where an individuals' protected attributes $A$ belonged to a different group.}
\end{definition}

\section{Methodology}\label{S-methods}
To investigate accuracy biases in contemporary ASR systems against disfluent speech, we have designed two experiments, each focusing on a different speech dataset. The first dataset is well-known in the speech-language pathology (SLP) community, and the second is widely used in ASR modeling and evaluation. Our objective is to compare ASR performance on two versions of these datasets: one with disfluencies and another identical in content but without disfluencies.

Experiment 1 utilizes the FluencyBank dataset \cite{ratner2018fluency}, known for its collection of real stuttered speech samples with accompanying transcripts (detailed in Sec. \ref{S-methods-dataset}). We refer to this original dataset as FluencyBank-Y (FB-Y), where ``Y'' indicates the presence of disfluencies. These samples are transcribed using six different ASR models (Sec. \ref{S-methods-models}), and the accuracy of these transcriptions is measured using WER, CER, and BERTScore, a metric for semantic similarity analysis (Sec. \ref{S-methods-metrics}). For comparative analysis, we generate a version of this dataset with apparent instances of stuttering or disfluencies removed, named FluencyBank-N (FB-N), where ``N'' signifies the absence of disfluencies, using text-to-speech (TTS) techniques (Sec. \ref{S-methods-generation}). We then employ the Spearman correlation coefficient test to systematically compare the performance of the ASRs on both the original FB-Y and the modified FB-N datasets.

To complement Experiment 1, Experiment 2 utilizes the LibriSpeech dataset \cite{panayotov2015librispeech}, frequently used in training and evaluating ASRs and characterized by standard, non-disfluent speech samples. We refer to this dataset as LibriSpeech-N or LS-N to convey the absence of disfluencies. To assess the impact of disfluencies, we create a synthetic dataset by introducing controlled disfluency events into the LibriSpeech dataset. This is achieved through a combination of audio processing and TTS methods. This newly modified dataset, denoted LibriSpeech-Y or LS-Y, where ``Y'' indicates the presence of disfluencies, is then subjected to the same testing process as the FluencyBank dataset in Experiment 1, using the same set of six ASR models. This approach ensures a systematic analysis of bias within these ASR systems across diverse speaking scenarios.

\subsection{ASR Models Studied}\label{S-methods-models}
We investigate a mix of commercially-available market ASRs and research-oriented, open-source models, to uncover biases across diverse ASR systems. The market ASRs include: (a) \textit{Google Cloud’s Speech-to-Text V1} \cite{gcp}, (b) \textit{IBM Watson’s Speech-to-Text} \cite{ibm}, (c) \textit{Microsoft Azure} \cite{azure}, and (d) \textit{RevAI} \cite{revai}. It is important to note that the specific details regarding the training processes of these commercial models are proprietary and not publicly disclosed.

In addition to these market ASRs, our investigation extends to two open-source, research-based ASR models, which are:

\begin{itemize}

    \item \textit{wav2vec 2.0} \cite{baevski2020wav2vec}:
    This is an end-to-end ASR model, utilizing connectionist temporal classification loss for training. For our analysis, we employ the `wav2vec2-large-960h' configuration, available from Meta on the HuggingFace Library \cite{wav2vecweights}. This specific model was trained on a substantial corpus of 960 hours of the LibriSpeech dataset.

    \item \textit{Whisper} \cite{radford2023robust}:
    Representing the current state-of-the-art in ASR technology, Whisper is a transformer-based encoder-decoder model. It has been pretrained on over 680k hours of multilingual data, leveraging semi-supervised learning and multitasking approaches—including tasks like language identification and voice activity detection. The primary aim of Whisper is to achieve generalization across a variety of benchmarks. For our purposes, we utilize the `whisper-1' model configuration provided by OpenAI API, which is the most up-to-date version at the time of this writing.

\end{itemize}

\begin{figure}[t]
\centering
\includegraphics[width=.48\columnwidth,trim={0cm 0cm 0cm 0cm},clip]{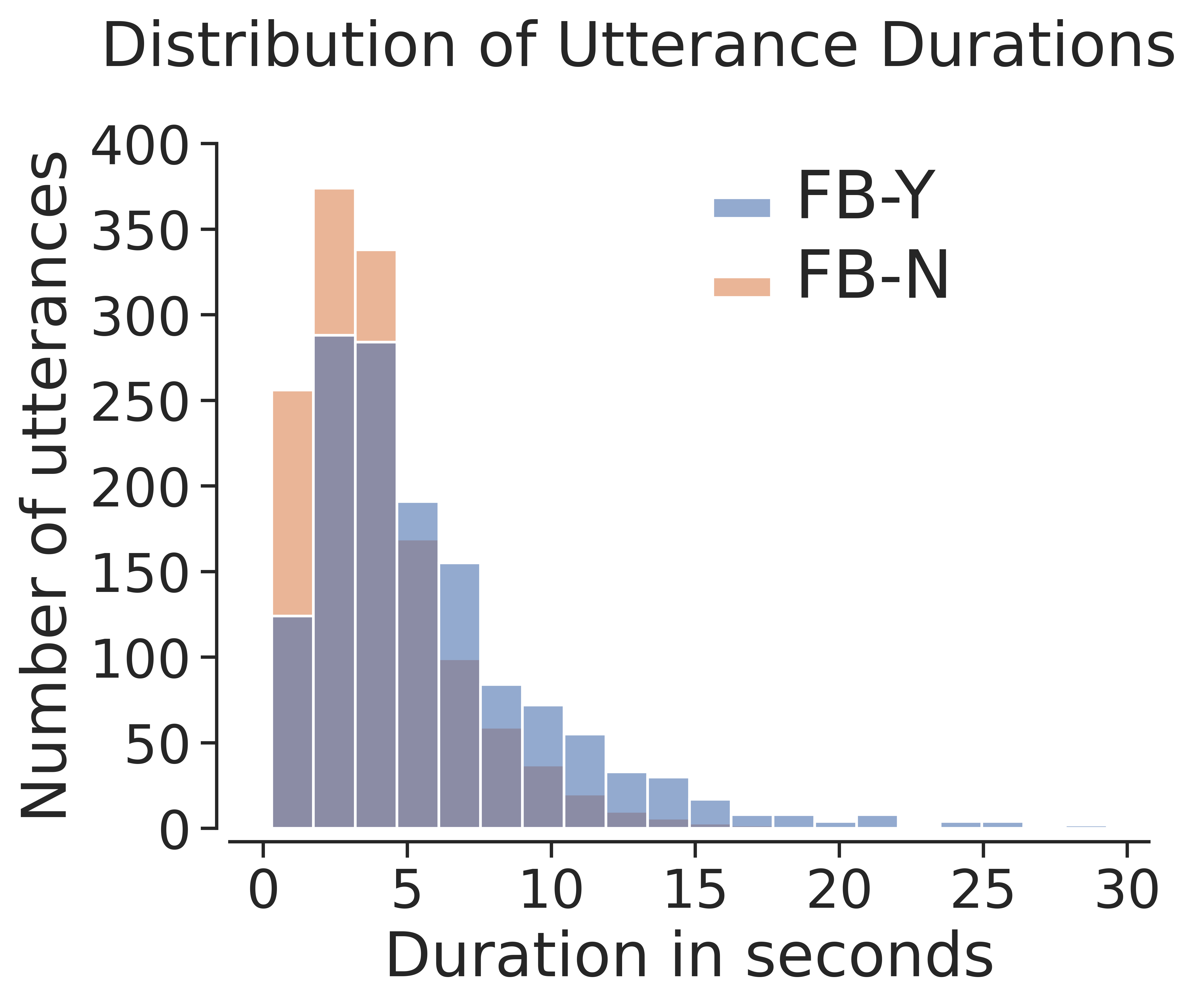}
\includegraphics[width=.49\columnwidth,trim={0cm 0cm 0cm 0cm},clip]{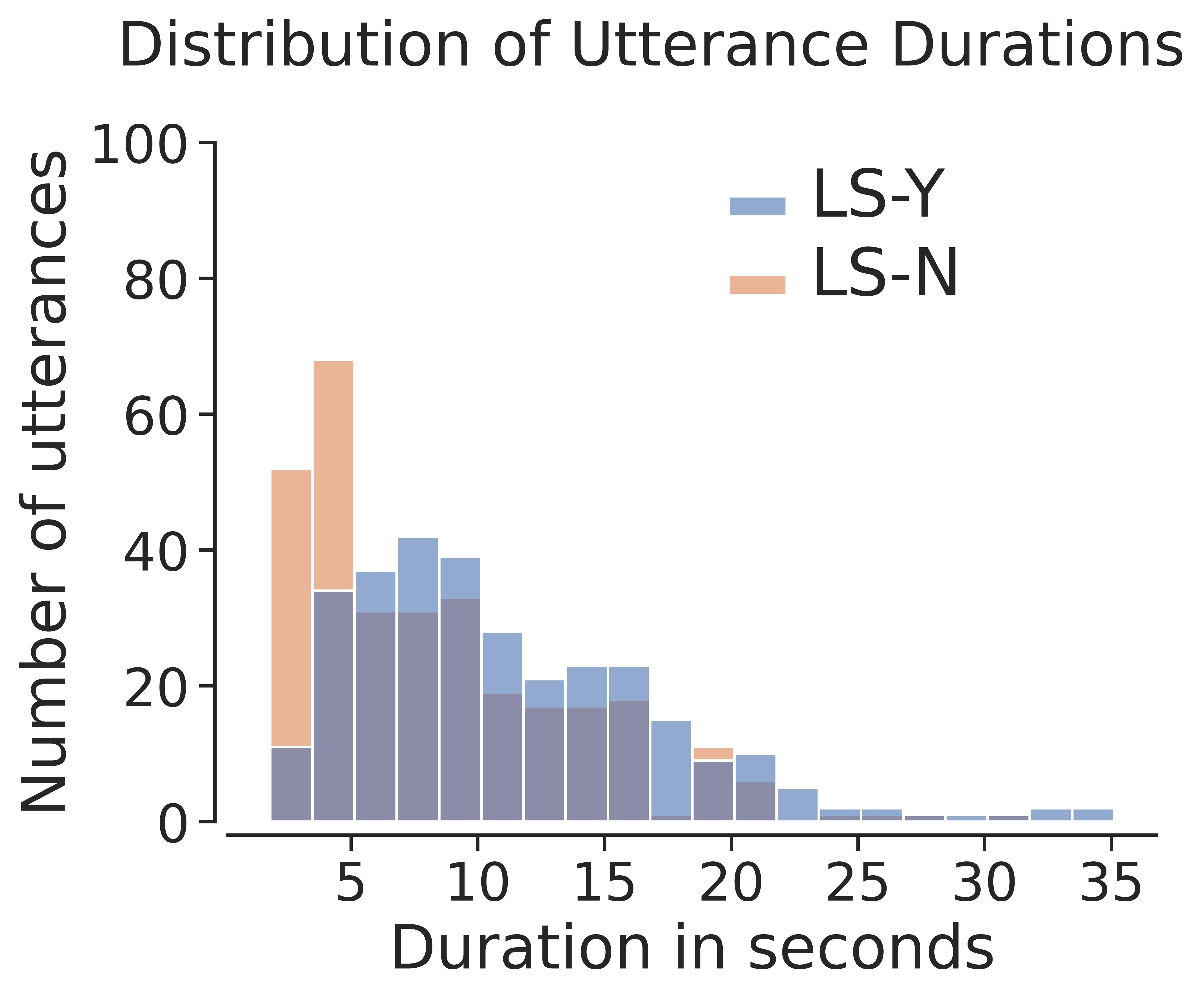}
\caption{Histogram showing the frequency distribution of utterance durations in seconds for each of the four speech datasets.}
\vspace*{-.15in}
\label{fig:datadist}
\end{figure}

\subsection{FluencyBank: Stuttered Speech Dataset}\label{S-methods-dataset}

Our study utilizes the FluencyBank dataset \cite{ratner2018fluency} to examine biases in real stuttered speech. FluencyBank comprises a series of video recordings featuring individuals who stutter, captured while reading passages and during interviews. Each video in this dataset is accompanied by a transcript, which has been segmented into individual utterances---a sentence or a fragment of a sentence (containing whole phrases) spoken by a single individual.

Due to temporal misalignment and incorrect transcript errors, which were identified by us and others \cite{sheikh2022machine,lea2021sep}, we had the dataset relabeled by trained SLPs with expertise in stuttering. They used the standard Codes for the Human Analysis of Transcripts format \cite{macwhinney2017tools}, and added start and end timestamps for each utterance within a given audio file, while ensuring inter-annotator agreement. The dataset utilized in our analysis consists of 1,373 utterances, derived from 7 reading and 12 interview videos. The distribution of these utterances in terms of duration is shown in Fig. \ref{fig:datadist}, with the cumulative audio duration amounting to approximately 2.21 hours.

In addition to the original dataset (FB-Y), we produced a version with apparent instances of stuttering or disfluencies removed, referred to as FluencyBank-N or FB-N. This variant was created using TTS, which synthesized speech from reference transcripts devoid of disfluencies. This approach, detailed in Sec. \ref{S-methods-generation}, serves as a crucial component in our comparative analysis of bias.

\subsection{Synthetic Disfluent Speech Generation}\label{S-methods-generation}

To complement our analysis using the FluencyBank dataset, we produce a synthetic dataset (LibriSpeech-Y or LS-Y) derived from LibriSpeech, an ASR dataset featuring public domain audiobook recordings. We utilized the `clean' sample from LibriSpeech's test set, available through the HuggingFace library, to ensure that the wav2vec 2.0 model was not previously exposed to this data during training. However, for commercial models (i.e., those available from Google and Microsoft), we do not have specific training data details and hence can not exclude the possibility that LibriSpeech has been used in their training, thus potentially boosting their performance in our analysis.

Our approach improves upon the LibriStutter dataset by FluentNet \cite{kourkounakis2020fluentnet}, which lacks diversity in disfluency events. LibriStutter's limitations include inadequate representation of the variability of stuttering, such as inserting only one to two word repetitions. Moreover, LibriStutter only uses audio processing methods for disfluency insertion. We introduce a more diverse range of disfluency events \cite{einarsdottir2005have}: \textit{interword blocks, intraword blocks, prolongations, word repetitions, phrase repetitions,} and \textit{interjections}. While these do not cover all types of disfluencies associated with stuttering, they serve to demonstrate key variations for our bias study.

We introduce disfluencies into the speech samples using two methods: \textit{textual disfluency insertion} and \textit{audio disfluency insertion}. For textual disfluency insertion, we modify the transcript and then use a TTS model, specifically SpeechT5 TTS \cite{ao-etal-2022-speecht5,ao2021speecht5}, to generate speech that naturally includes the disfluency events. This process, supported by a set of pre-trained speaker vectors \cite{cmuvectors}, allows us to simulate a diverse range of voices. This technique is primarily used for word repetitions, phrase repetitions, and interjections.
The frequency and location of these disfluencies within each sentence are selected randomly within predetermined ranges: (1) for word repetition, we select between 1 to 3 words in each utterance to repeat 1 to 4 additional times; (2) for phrase repetition, we choose a phrase of 2 to 4 words and repeat it 1 to 3 times; (3) for interjections, we randomly insert either ``uh'' or ``um'' in 1 to 4 locations and repeat them 1 to 4 times.

For audio disfluency insertion, we directly modify the speech waveform to add disfluencies that TTS cannot create, such as blocks and prolongations. We employ a pre-trained wav2vec 2.0 model and PyTorch CTC forced alignment \cite{alignmentpytorch} for precise alignment of words with their timestamps. This method allows for the insertion of interword blocks, intraword blocks, and additional word and phrase repetitions with parameters similar to those used in textual disfluency insertion. The parameters for disfluency events applied in audio processing are also randomly selected: (1) for interword blocks, we insert a pause after 1 to 4 words with a duration of 1 to 3 seconds; (2) for intraword blocks, we insert a pause into 1 to 3 words, each lasting between 1 to 3 seconds; (3) for prolongations, we extend the sound of 1 to 3 words at their beginning, stretching it to 5 to 25 times its original length. Additionally, word and phrase repetitions are applied following the same parameters as in the textual disfluency insertion process.

Each disfluency is randomly applied, without replacement, to an equal number of samples, amounting to 10\% of the LibriSpeech dataset, ensuring a balanced representation. The resulting dataset comprises approximately one hour of speech samples, with 308 records in total. The duration of these samples, in seconds, is depicted in Figure \ref{fig:datadist}.

\begin{table}[t]
  \begin{center}
  \caption{Examples of disfluency events including word repetition (WR), phrase repetition (PR), prolongation (Prolong.), interword block (InterWB), intraword block (IntraWB), and interjection (Interject.). Words in <> brackets are added through textual disfluency insertion, while (pause) signifies the audio insertion of silence.}
  \begin{tabular}{ p{2.7cm}p{4cm} } 
  \hline
  \textbf{Disfluency Event} & \textbf{Example}  \\ \hline
    WR & \textit{How <are> are you?}  \\
    PR & \textit{How <are you> are you?}\\
    Interject. & \textit{How are <um> you?}\\
    \hline
    InterWB& \textit{How are (pause) you?}\\
    IntraWB& \textit{How a(pause)re you?} \\
    Prolong.& \textit{How aaare you?} \\
  \hline
\end{tabular}
\vspace*{-.2in}
\label{tbl:stutterevents}
\end{center}
\end{table}

This approach allows for a comprehensive and nuanced analysis of ASR performance across a range of disfluency types, thereby enriching our understanding of bias in these systems. 

\subsection{ASR Evaluation Metrics}\label{S-methods-metrics}
To effectively assess biases in ASR systems, we employ three key metrics, each offering a unique lens on transcription accuracy and semantic fidelity.

The first metric is the \textit{word error rate (WER)}, a widely recognized standard in ASR evaluations, defined as:
$
{\rm WER} = \frac{S + D + I}{S + D + C},
$
where $S$, $D$, $I$, and $C$ represent the number of word substitutions, word deletions, word insertions, and correct words, respectively, in the ASR-produced transcript relative to the reference transcript. WER provides a direct measure of transcription accuracy at the word level. A lower WER, approaching zero, is desirable, indicating higher word transcription accuracy.

In conjunction with WER, we analyze the \textit{character error rate (CER)}, which evaluates transcription errors at the character level. Similar to WER but more granular, CER offers insight into the accuracy of finer transcription details. As with WER, a lower CER is indicative of better performance.

Our third metric, \textit{BERTScore} \cite{zhang2019bertscore}, examines semantic similarity between predicted and target sentences. This metric is crucial for cases where a transcription error might involve just a single word, yet lead to a significant change in meaning. For instance, incorrectly transcribing "not" in a sentence could completely reverse its intended meaning, despite being a single-word error. BERTScore assesses this aspect, with a score closer to 1 indicating greater semantic similarity between the predicted and target sentences.

With BERTScore, we can observe the meaning behind each word. Given an ASR-produced sentence $x = \left \langle x_1, ..., x_k \right \rangle$ and a reference sentence $\hat{x} = \left \langle \hat{x}_1, ..., \hat{x}_l \right \rangle$, where $x_i$ and $\hat{x}_i$ represent the $i^{th}$ token embeddings derived directly from a pre-trained BERT model's contextual embeddings, BERTScore computes the cosine similarity of each token in $x$ with every token in $\hat{x}$ to obtain an F1 score. We utilize embeddings from the `microsoft/deberta-large-mnli' model \cite{huggingface_deberta_large_mnli} alongside the re-scaled version of BERTScore, wherein the range is within $[-1, 1]$, where a 1 would be semantically identical transcripts, and -1 opposite. We refer to this as $F_{BERT}$ \cite{tobin2022assessing,zhang2019bertscore} and it is defined as the following:
\begin{equation}
F_{BERT} = 2\frac{P_{BERT}\times R_{BERT}}{P_{BERT} + R_{BERT}}, {\textrm{where}}
\end{equation}
\begin{equation}
R_{BERT} = \frac{1}{|x|}\sum_{x_i\in x}\max_{\hat{x}_j \in \hat{x}}(x_i^\textup{T} \cdot \hat{x}_j), {\textrm{and}}
\end{equation}
\begin{equation}
P_{BERT} = \frac{1}{|\hat{x}|}\sum_{\hat{x}_i\in \hat{x}}\max_{x_j \in x}(x_i^\textup{T} \cdot \hat{x}_j).
\end{equation}

We use the Spearman correlation coefficient to analyze the relationship between ASR performance metrics across disfluent and non-disfluent speech datasets. We opt for this non-parametric test because the performance distributions are non-normal (as detailed in Appendix \ref{appendix:spearman-results}), necessitating an analysis approach that does not assume normality. The Spearman correlation coefficient, defined as:
\begin{equation}
r_{WER}= 1 - \frac{6\sum d_i^2}{n (n^2 - 1)},
\end{equation}
facilitates our understanding of variations in ASR model performance across the two datasets. Here, $d_i$ is the difference in ranks between each observation's WER in disfluent and non-disfluent speech datasets, and $n$ represents the total number of observations or utterances. This approach helps us quantify the impact of disfluent speech patterns, like stuttering, on ASR accuracy. A value closer to 1 or -1 indicates a strong positive or negative correlation, respectively, between the presence of disfluencies and the difference in WER, suggesting a significant impact of disfluencies on ASR performance. Conversely, a correlation closer to 0 implies a weaker relationship. We perform similar computations for CER ($r_{CER}$) and BERTScore ($r_{BERT}$), offering a multidimensional view of each ASR system's bias and accuracy.

Additionally, we incorporate the calculation of the \textit{p-value} to determine the statistical significance of the observed biases. The $p$-value quantifies the probability of observing the results under the null hypothesis, which in this context is the absence of bias. A $p$-value lower than a typical threshold (e.g., 0.05) suggests that the observed bias is statistically significant and not due to random chance. This metric is essential for substantiating the reliability of our findings.

\section{Results and Discussion}\label{S-results}
Our analysis, divided into two parts corresponding to each experiment in Sec. \ref{S-methods}, reveals significant insights into ASR performance with real and synthetic disfluent speech. The evaluation was conducted using ASR services as of March. 14, 2024, utilizing default API settings. For wav2vec 2.0, the analysis was performed locally using an NVIDIA T4 GPU. Transcripts returned blank by models due to audio processing challenges (e.g., due to noise or blocks) were retained in our dataset. Prior to evaluation, all transcripts underwent normalization using Whisper's ``BasicTextNormalizer'' and ``EnglishNumberNormalizer'' functions from the Transformers library \cite{normalizer}, ensuring consistent casing and punctuation removal. Any speech atmospherics in output transcripts, such as laughing or indications of silence, were also removed.

\subsection{Experiment 1: Real Stuttered Speech}\label{S-results-1}

\begin{table}[t]
  \begin{center}
  \caption{Comparative analysis of ASR model performance metrics (WER, CER, and $F_{BERT}$) on FluencyBank datasets: `Y' denotes the presence of stuttering (FB-Y), and `N' denotes its absence (FB-N). The mean of scores across ASRs is given by $\mu$, and the standard deviation by $\sigma$.}
  \begin{tabular}{ p{1cm}|p{0.6cm}p{0.6cm}|p{0.6cm}p{0.6cm}|p{0.6cm}p{0.6cm} } 
  \toprule
    & \multicolumn{2}{c|}{WER} & \multicolumn{2}{c|}{CER} & \multicolumn{2}{c}{$F_{BERT}$} \\
   & \multicolumn{1}{c}{Y} &\multicolumn{1}{c|}{N}& \multicolumn{1}{c}{Y}&\multicolumn{1}{c|}{N}& \multicolumn{1}{c}{Y}&\multicolumn{1}{c}{N}\\\midrule
Azure & .168 & \textbf{.073}    & .117 & \textbf{.052} & .805 & \textbf{.896} \\
GCP & .275 & \textbf{.069}      & .182 & \textbf{.050} & .623 & \textbf{.920} \\
IBM & .476 & \textbf{.167}      & .407 & \textbf{.150} & .463 & \textbf{.838} \\
RevAI & .263 & \textbf{.159}    & .187 & \textbf{.133} & \textbf{.586} & {.508} \\
Whisp  & .121 & \textbf{.063}    & .082 & \textbf{.047} & .870 & \textbf{.935} \\
W2V2 & .380 & \textbf{.101}     & .212 & \textbf{.061} & .549 & \textbf{.865} \\
\midrule
$\mu$ & .281 & \textbf{.105}     & .198 & \textbf{.082} & .649 & \textbf{.827} \\
$\sigma$ & .120 & {.042} & .103 & {.042} & .143 & {.146} \\
  \bottomrule
\end{tabular}
\vspace*{-.2in}
\label{tbl:e1metrics}
\end{center}
\end{table}

\begin{table}[t]
  \begin{center}
  \caption{Spearman correlation coefficients ($r$), all of which had $p < 0.05$, indicating the statistical significance of the impact of disfluencies on $\bigtriangleup$WER, $\bigtriangleup$CER, and $\bigtriangleup$$F_{BERT}$. FB refers to comparisons between FB-Y and FB-N datasets, and LS to comparisons between synthetic (LS-Y) and original (LS-N) LibriSpeech datasets. ``All'' refers to the combination of all ASR predictions.}
  \begin{tabular}{ p{1cm}|p{0.6cm}p{0.6cm}|p{0.6cm}p{0.6cm}|p{0.63cm}p{0.63cm} } 
  \toprule
    & \multicolumn{2}{c|}{$\bigtriangleup$WER} & \multicolumn{2}{c|}{$\bigtriangleup$CER} & \multicolumn{2}{c}{$\bigtriangleup$$F_{BERT}$} \\
   & \multicolumn{1}{c}{FB} &\multicolumn{1}{c|}{LS}& \multicolumn{1}{c}{FB}&\multicolumn{1}{c|}{LS}& \multicolumn{1}{c}{FB}&\multicolumn{1}{c}{LS}\\\midrule
Azure & .363 & .656    & .385 & .700 & -.33 & -.62 \\
GCP & .552 & .547      & .559 & .609 & -.35 & -.47 \\
IBM & .531 & .616      & .539 & .690 & -.52 & -.49 \\
RevAI & .245 & .557   & .246 & .611 & -.21 & -.50 \\
Whisp  & .285 & .501   & .298 & .528 & -.24 & -.48 \\
W2V2 & .563 & .712     & .562 & .781 & -.54 & -.67 \\
\midrule
All & .413 & .566     & .423 & .623 & -.38 & -.50 \\
  \bottomrule
\end{tabular}
\vspace*{-.2in}
\label{tbl:pvalue}
\end{center}
\end{table}

The outcomes of ASR model performance on the FluencyBank dataset, with and without disfluencies (denoted as FB-Y and FB-N, respectively), are presented in Table \ref{tbl:e1metrics}. Our findings indicate a consistent pattern across all models: lower WER, CER, and higher $F_{BERT}$ on the FB-N dataset, implying a systemic bias against disfluent speech. Among the evaluated models, Whisper exhibited the lowest WER and highest $F_{BERT}$. However, its performance on FB-Y was markedly worse (about two times) compared to that on FB-N, where it achieved a WER of just 6.3\%. We also note that the RevAI model showed a higher $F_{BERT}$ for FB-Y than for FB-N, primarily due to blank transcript predictions. Upon removing these from the analysis, the reported $F_{BERT}$ on FB-Y is 0.809 and on FB-N is 0.933. This disparity in performance highlights a significant bias in ASRs against disfluent speech.

\begin{figure}[t!]
    \centering
	\includegraphics[width=.75\columnwidth,trim={1.4cm 0cm 1.5cm 0cm},clip]{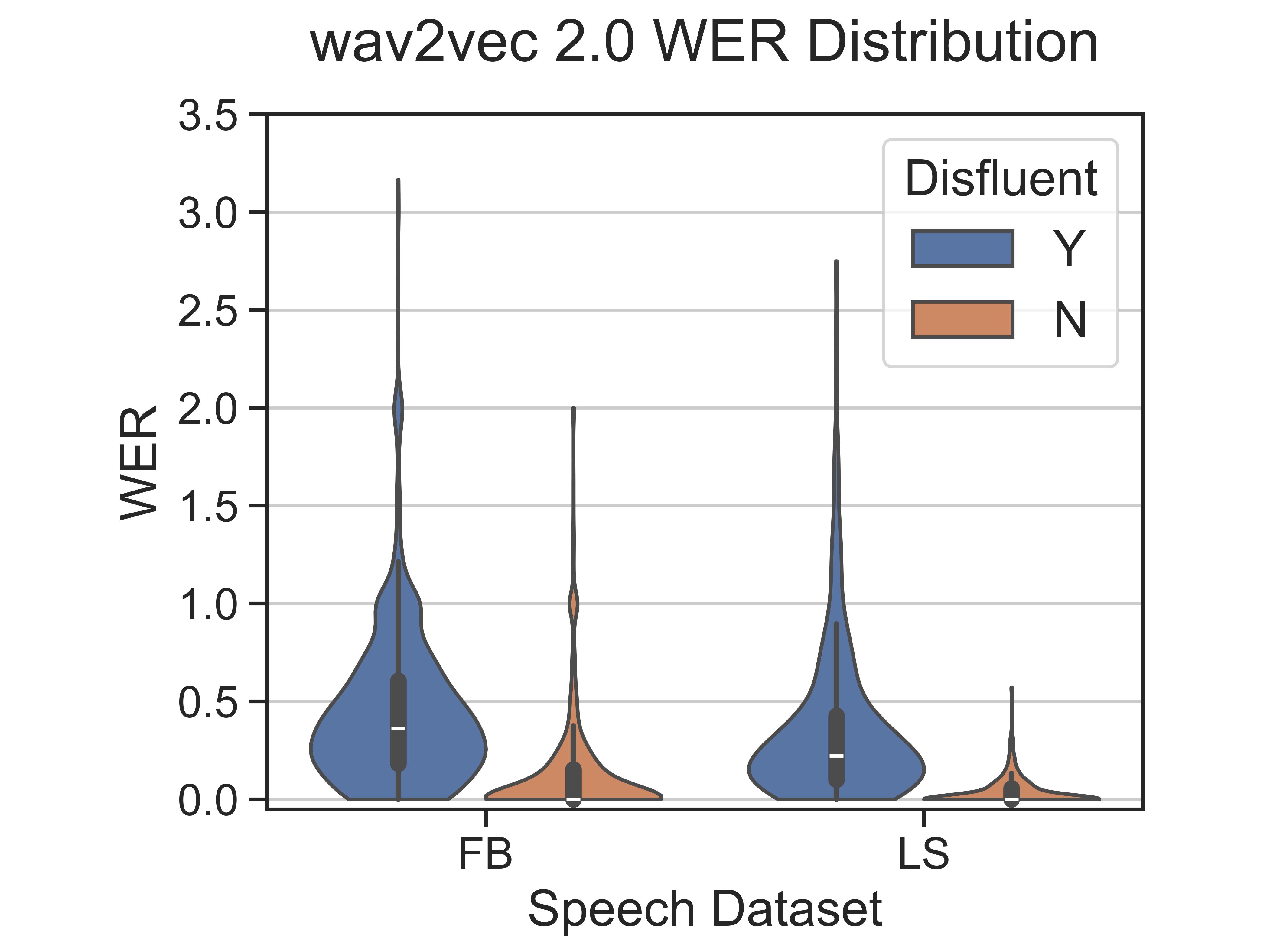}
  \vspace*{-.1in}
    \caption{Violin plot illustrating the range and distribution of WER for the wav2vec 2.0 model on FluencyBank (FB) and synthetic (LS) datasets. The plot highlights mean values and variability in WER scores.}
	\label{fig:distplot}
   \vspace*{-.2in}
\end{figure}

Moreover, Fig. \ref{fig:distplot} illustrates the WER distribution for the wav2vec 2.0 model across all datasets. The mean WER is noticeably higher for datasets with disfluencies, highlighting a broader range and more outliers with elevated WER.

\subsection{Experiment 2: Synthetic Speech}\label{S-results-2}
In our second experiment, we evaluated the ASR models against our synthetic disfluency dataset compared to the original LibriSpeech. The results (Table \ref{tbl:e2metrics}) mirror those from Experiment 1, with higher WER and CER for the synthetic disfluency dataset. This finding confirms the bias observed in Experiment 1, not just for real disfluent speech but also for synthesized versions. The Whisper model again showed superior performance, but with a significant drop (WER of 3.1\% vs. 19.8\%) when disfluencies were introduced.

\begin{table}[t]
  \begin{center}
  \caption{Evaluation of ASR models on synthetic stuttering and original LibriSpeech datasets. Metrics include WER, CER, and $F_{BERT}$, with `Y' indicating synthetic stuttering (LS-Y) and `N' indicating the original dataset (LS-N). The mean of scores across ASRs is given by $\mu$, and the standard deviation by $\sigma$.}
  \begin{tabular}{ p{1cm}|p{0.6cm}p{0.6cm}|p{0.6cm}p{0.6cm}|p{0.6cm}p{0.6cm} } 
  \toprule
    & \multicolumn{2}{c|}{WER} & \multicolumn{2}{c|}{CER} & \multicolumn{2}{c}{$F_{BERT}$} \\
   & \multicolumn{1}{c}{Y} &\multicolumn{1}{c|}{N}& \multicolumn{1}{c}{Y}&\multicolumn{1}{c|}{N}& \multicolumn{1}{c}{Y}&\multicolumn{1}{c}{N}\\\midrule
Azure & .228  & \textbf{.037}   & .197 & \textbf{.018}  & .772 & \textbf{.957} \\
GCP & .268 & \textbf{.070}      & .217 & \textbf{.034}  & .744 & \textbf{.904} \\
IBM & .560 & \textbf{.191}      & .518 & \textbf{.127}  & .448 & \textbf{.757} \\
RevAI & .254 & \textbf{.062}      & .209 & \textbf{.031}  & .742 & \textbf{.901} \\
Whisp & .198 & \textbf{.031}    & .157 & \textbf{.015}  & .811 & \textbf{.969} \\
W2V2 & .259 & \textbf{.033}     & .219 & \textbf{.009}  & .718 & \textbf{.952} \\
\midrule
$\mu$ & .295 & \textbf{.071}    & .253 & \textbf{.039}   & .706 & \textbf{.907} \\
$\sigma$ & .121 & .056 & .120 & .040  & .119 & .072 \\
  \bottomrule
\end{tabular}
\vspace*{-.2in}
\label{tbl:e2metrics}
\end{center}
\end{table}

\begin{table*}[t]
  \begin{center}
  \caption{WER results for each ASR on the synthetic dataset, broken down by specific disfluency event types. These include audio-based word repetition (Aud-WR), audio-based phrase repetition (Aud-PR), prolongation (Prolong.), interword block (InterWB), intraword block (IntraWB), interjection (Interject.), text-TTS-based word repetition (Txt-WR), and text-TTS-based phrase repetition (Txt-PR). `Y' denotes the presence of each disfluency event type, and `N' indicates its absence. The mean of scores across ASRs is given by $\mu$, and the standard deviation by $\sigma$.}
  \begin{tabular}{ l|c >{\bfseries}c|c >{\bfseries}c|c >{\bfseries}c|c >{\bfseries}c|c >{\bfseries}c|c >{\bfseries}c|c >{\bfseries}c|c >{\bfseries}c } 
  \toprule
     & \multicolumn{2}{c|}{Aud-WR} & \multicolumn{2}{c|}{Aud-PR} &\multicolumn{2}{c|}{Prolong.}  & \multicolumn{2}{c|}{InterWB} & \multicolumn{2}{c|}{IntraWB} & \multicolumn{2}{c|}{Interject.} & \multicolumn{2}{c|}{Txt-WR} & \multicolumn{2}{c}{Txt-PR} \\
    &Y&\normalfont{N}&Y&\normalfont{N}&Y&\normalfont{N}&Y&\normalfont{N}&Y&\normalfont{N}&Y&\normalfont{N}&Y&\normalfont{N}&Y&\normalfont{N} \\
    \midrule
Azure	&.27	&.02	&.22	&.03	&.35	&.04	&.12	&.05	&.08	&.03	&.32	&.05	&.32	&.05	&.26	&.03\\
GCP	&.33	&.05	&.24	&.06	&.53	&.08	&.11	&.08	&.08	&.05	&.38	&.07	&.36	&.09	&.28	&.10\\
IBM	&.67	&.42	&.35	&.22	&.65	&.20	&.72	&.21	&.68	&.11	&.52	&.15	&.42	&.20	&.49	&.12\\
RevAI	&.34	&.05	&.23	&.04	&.44	&.08	&.10	&.08	&.07	&.05	&.36	&.08	&.36	&.09	&.28	&.04\\
Whisp	&.23	&.03	&.12	&.03	&.32	&.03	&.07	&.04	&.04	&.02	&.41	&.04	&.33	&.05	&.23	&.02\\
W2V2	&.33	&.02	&.21	&.03	&.55	&.03	&.11	&.03	&.09	&.03	&.34	&.04	&.32	&.04	&.26	&.05\\
\midrule
$\mu$	&.36	&.10	&.23	&.07	&.47	&.08	&.20	&.08	&.17	&.05	&.39	&.07	&.35	&.09	&.30	&.06\\
$\sigma$	&.14	&\normalfont .14	&.07	&\normalfont .07	&.12	&\normalfont .06	&.23	&\normalfont .06	&.23	&\normalfont .03	&.07	&\normalfont .04	&.04	&\normalfont .05	&.08	& \normalfont .04\\
    \bottomrule
  \end{tabular}
  \vspace*{-.2in}
  \label{tbl:e2eventmetrics}
  \end{center}
\end{table*}

Further analysis on disfluency-event-type specific WER (Table \ref{tbl:e2eventmetrics}) reveals that certain disfluencies, such as word repetitions, prolongations, and interjections, significantly increase WER. In contrast, intraword block and interword block events showed a less pronounced impact. This suggests that different disfluency types vary in their effect on ASR performance, underscoring the need for more inclusive and diverse training data in ASRs.

The Spearman correlation coefficients and corresponding $p$-values (Table \ref{tbl:pvalue}) across all models and for different disfluencies (Appendix \ref{appendix:spearman-disfluencies}) underscore this finding. We find all correlation values in Table \ref{tbl:pvalue} are statistically significant, with $p < 0.05$, indicating that the presence of disfluencies substantially impacts ASR accuracy. For example, prolongations and interjections showed a high correlation with increased WER, highlighting specific areas where ASRs struggle in handling disfluent speech. We also note the correlation coefficients for $F_{BERT}$ present an inverse correlation, where ASRs consistently had lower $F_{BERT}$ scores on FB-Y and LS-Y, indicating poor semantic similarity.  

Overall, our results indicate a clear bias in current ASR technology against disfluent speech, both real and synthetic. This bias is evident across various disfluency event types and is statistically significant, emphasizing the need for targeted improvements in ASRs to accommodate speech variations.

\section{Conclusion}\label{S-conclusion}
This study represents a significant step in identifying and quantifying the biases in ASR systems against individuals who stutter. By analyzing the performance of six popular ASR models using two distinct datasets—FluencyBank and a custom-generated synthetic dataset incorporating disfluency events from LibriSpeech—we have uncovered substantial biases. Our findings reveal that all six models demonstrate notably worse WER, CER, and BERTScore when processing disfluent speech. This bias is consistent across both real and synthetic datasets and is statistically significant.

Our research also delves into disfluency-event-specific biases, revealing that the majority of disfluencies—six out of eight types—exhibit a statistically significant WER bias across all models. This discovery is crucial in understanding the nuances of ASR performance and the challenges faced by people who stutter in real-world scenarios.

To address these challenges, there are a few approaches that can be considered. First, incorporating speech from people who stutter in ASR training; this is challenging due to the limited availability of relevant data. Additionally challenging is the task of acquiring sufficient data that encompasses samples from a diverse population to capture the variability in speech associated with stuttering. However, with a strategic data collection approach, this can address a primary limitation in ASR accessibility. Secondly, obtaining feedback from people who stutter and benchmarking ASRs before deployment are crucial steps towards accessibility. Rigorous testing on disfluent speech can identify ASR deficiencies, allowing for targeted training or post-processing to address them.

Looking forward, we also identify two promising directions for future research. Firstly, there is a need to explore the impact of longer-form audio on ASR biases. Current models often struggle with correctly segmenting and transcribing longer audio samples containing disfluencies, leading to significant transcription inaccuracies or even cessation of live transcription. This aspect is particularly critical as it can severely limit the practical usability of ASR technologies for individuals who stutter.

Secondly, incorporating human feedback from people who stutter represents an important area of exploration. Current metrics like WER and semantic accuracy might not fully capture the transcription preferences of individuals who stutter. While our study assumes that the transcription of FluencyBank, with stuttering indicators removed, accurately represents the intended speech, this may not align with the preferences of those who stutter. They might desire their speech patterns, such as interjections, prolongations, and other disfluencies, to be reflected in the transcripts. Hence, engaging directly with individuals who stutter to establish ground-truth labels could offer a more authentic and respectful approach to ASR development.

In conclusion, our research is among the first to systematically examine and highlight the biases in ASR systems towards disfluent speech. The implications of these biases are far-reaching, potentially exacerbating disparities in various domains, from healthcare to recruitment, where ASR technologies are increasingly integrated. To pave the way for more inclusive and equitable ASR technologies, bias mitigation strategies, including diversifying training data with disfluent speech and incorporating feedback from the stuttering community, are essential. Our work lays the foundation for this crucial endeavor, opening avenues for future research and development in the field of speech recognition.

\section*{Limitations}
We acknowledge three limitations of our work. First, our study focuses solely on accuracy bias by ASR models transcribing speech in English, and people who stutter speaking English. This does not cover the many multilingual ASR models and people who stutter speaking other languages. Second, our synthetic dataset does not encompass all types of disfluencies, and not all disfluencies (e.g., interjections) can be considered stuttering in all circumstances since they depend upon the presence of a loss of control by the speaker while speaking \cite{tichenor2019stuttering}. Our intended use for this dataset is for studying bias at a fine-grained disfluency level, with control over the frequency and location of occurrence of disfluencies. Therefore, this dataset should not be used for ASR model training as a bias mitigation measure, and should also not be used to study bias unaccompanied by other datasets. Likewise, third, our study method should not be used as the sole assessment of bias in ASRs. We define bias by observing the transcript accuracy (i.e., WER, CER, and semantic accuracy) difference with and without disfluencies. However, other metrics for bias and other facets, such as age or gender, can affect ASR bias. Overall, we seek to develop more inclusive ASRs, and have these limitations addressed in future work.

\section*{Acknowledgment}
This material is based upon work supported by the U.S. National Science Foundation under Grant Nos. 2235916 and 2345086.

\bibliography{refs}

\appendix

\begin{table}[t!]
  \begin{center}
  \caption{Comparative analysis of similarity scores between our synthetic dataset LS-Y with disfluencies, and both FluencyBank datasets where `Y' denotes the presence of disfluencies (FB-Y), and `N' denotes its absence (FB-N). This is broken down by specific disfluency event types. These include audio-based word repetition (Aud-WR), audio-based phrase repetition (Aud-PR), prolongation (Prolong.), interword block (InterWB), intraword block (IntraWB), interjection (Interject.), text-TTS-based word repetition (Txt-WR), and text-TTS-based phrase repetition (Txt-PR). The mean of similarity scores is given by $\mu$, and the standard deviation by $\sigma$. ``All' is the average distance across all disfluency types. We observe across all disfluency types, our synthetic speech dataset is closer to resembling stuttered speech than non-stuttered speech.}
  \begin{tabular}{ p{1.6cm}|p{0.75cm}p{0.75cm}|p{0.75cm}p{0.75cm} } 
  \toprule
    & \multicolumn{2}{c|}{FB-Y} & \multicolumn{2}{c}{FB-N}  \\
   & \multicolumn{1}{c}{$\mu$} &\multicolumn{1}{c|}{$\sigma$}& \multicolumn{1}{c}{$\mu$}&\multicolumn{1}{c}{$\sigma$}\\\midrule
Aud-WR & \textbf{.659} & .178    & .614 & .190  \\
Aud-PR & \textbf{.589} & .192    & .524 & .192  \\
Prolong. & \textbf{.634} & .178    & .590 & .190  \\
InterWB & \textbf{.637} & .187    & .569 & .192  \\
IntraWB & \textbf{.546} & .190    & .472 & .176  \\
Interject. & \textbf{.667} & .174    & .635 & .189  \\
Txt-WR & \textbf{.648} & .184    & .597 & .197  \\
Txt-PR & \textbf{.598} & .193    & .531 & .194  \\
\midrule
All & \textbf{.623} & .189    & .567 & .197  \\
  \bottomrule
\end{tabular}
\vspace*{-.1in}
\label{tbl:dataeval}
\end{center}
\end{table}

\begin{table*}[t!]
  \begin{center}
  \caption{Spearman correlation coefficient ($r_{WER}$ denoted by $r$ below) and corresponding $p$-value results for each ASR model on the synthetic dataset, broken down by specific disfluency event types. These include audio-based word repetition (Aud-WR), audio-based phrase repetition (Aud-PR), prolongation (Prolong.), interword block (InterWB), intraword block (IntraWB), interjection (Interject.), text-TTS-based word repetition (Txt-WR), and text-TTS-based phrase repetition (Txt-PR). `Y' denotes the presence of each disfluency event type, and `N' indicates its absence. ``All'' refers to the combination of all ASR predictions.}
  \begin{tabular}{ l|cc|cc|cc|cc|cc|cc|cc|cc } 
  \toprule
     & \multicolumn{2}{c|}{Aud-WR} & \multicolumn{2}{c|}{Aud-PR} &\multicolumn{2}{c|}{Prolong.}  & \multicolumn{2}{c|}{InterWB} & \multicolumn{2}{c|}{IntraWB} & \multicolumn{2}{c|}{Interject.} & \multicolumn{2}{c|}{Txt-WR} & \multicolumn{2}{c}{Txt-PR} \\
   & \multicolumn{1}{c}{$r$} &\multicolumn{1}{c|}{$p$}& \multicolumn{1}{c}{$r$} &\multicolumn{1}{c|}{$p$}& \multicolumn{1}{c}{$r$} &\multicolumn{1}{c|}{$p$}& \multicolumn{1}{c}{$r$} &\multicolumn{1}{c|}{$p$}& \multicolumn{1}{c}{$r$} &\multicolumn{1}{c|}{$p$}& \multicolumn{1}{c}{$r$} &\multicolumn{1}{c|}{$p$}& \multicolumn{1}{c}{$r$} &\multicolumn{1}{c|}{$p$}& \multicolumn{1}{c}{$r$} &\multicolumn{1}{c}{$p$}\\\midrule
Azure	&.78	&.00	&.83	&.00	&.40	&.00	&.56	&.00	&.58	&.00	&.70	&.00	&.76	&.00	&.79	&.00\\
GCP	&.72	&.00	&.81	&.00	&.45	&.00	&.18	&.11	&.27	&.01	&.71	&.00	&.69	&.00	&.62	&.00\\
IBM	&.54	&.00	&.51	&.00	&.48	&.00	&.76	&.00	&.84	&.00	&.61	&.00	&.48	&.00	&.66	&.00\\
RevAI	&.79	&.00	&.80	&.00	&.46	&.00	&.11	&.33	&.29	&.01	&.65	&.00	&.68	&.00	&.74	&.00\\
Whisp	&.64	&.00	&.40	&.00	&.36	&.00	&.17	&.13	&.10	&.40	&.77	&.00	&.74	&.00	&.78	&.00\\
W2V2	&.82	&.00	&.83	&.00	&.71	&.00	&.48	&.00	&.66	&.00	&.78	&.00	&.78	&.00	&.72	&.00\\
\midrule
All	&.67	&.00	&.67	&.00	&.46	&.00	&.33	&.00	&.40	&.00	&.69	&.00	&.67	&.00	&.70	&.00\\
    \bottomrule
  \end{tabular}
  \vspace*{-.1in}
  \label{tbl:stutterevents-detailed}
  \end{center}
\end{table*}

\section{Evaluation of Synthetic Disfluent Speech}\label{appendix:syneval}
To ensure that our synthetic disfluent speech dataset (LS-Y) presented in Section \ref{S-methods-generation} is comparable to real stuttered speech, we assess its similarity to the FluencyBank dataset (FB-Y). It is important to note that our dataset's objective is not to mimic stuttering precisely; instead, it aims to provide a comparable framework for evaluating ASR biases toward stuttered speech.

To compare, our evaluation procedure starts by computing the cosine similarity between all spectrograms of audio samples in LS-Y and FB-Y. Following this, we calculate the cosine similarity between all spectrograms of audio samples in LS-Y and FB-N, our corresponding FluencyBank dataset without disfluencies. This approach aims to determine whether our dataset exhibits greater similarity to FB-Y, which contains stuttered speech, than to FB-N, which lacks stuttering. Such a comparison helps gauge the resemblance of our dataset to stuttered speech versus non-stuttered speech.

The effectiveness of this approach and our dataset is demonstrated through the results presented in Table \ref{tbl:dataeval}, which display the average and standard deviation of similarities on a disfluency-specific level between LS-Y and both FB-Y and FB-N datasets. We observe that our synthetic speech dataset exhibits a greater similarity to real stuttered speech (FB-Y) than to non-stuttered speech (FB-N) across all disfluency types. This finding is further supported by the fact that our FB-N dataset was generated using voices through TTS (Sec. \ref{S-methods-generation}) overlapping with our LS-Y dataset, possibly increasing the similarity between LS-Y and FB-N. Therefore, overall, our synthetic stuttered speech generation method proves to be an effective means of introducing disfluencies in non-stuttered speech.

\section{Additional Experiment Results}\label{appendix:spearman}
\subsection{ASR performance distributions}\label{appendix:spearman-results}
\begin{figure*}[htp]
\centering
\includegraphics[width=.31\textwidth,trim={1.5cm 0cm 1.5cm 0cm},clip]{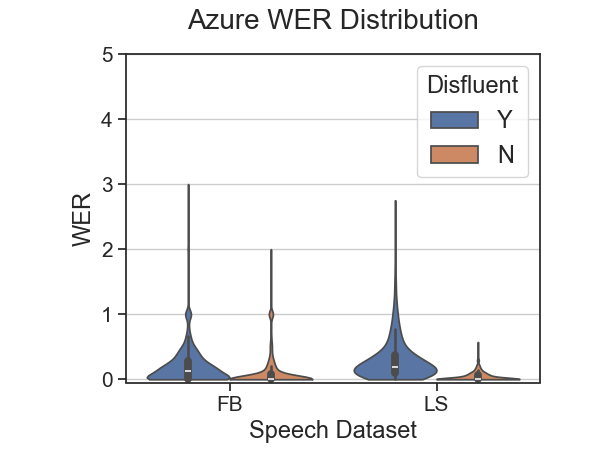}\quad
\includegraphics[width=.31\textwidth,trim={1.5cm 0cm 1.5cm 0cm},clip]{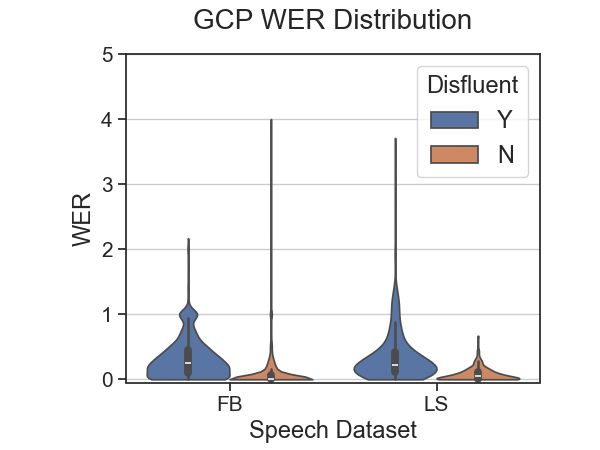}\quad
\includegraphics[width=.31\textwidth,trim={1.5cm 0cm 1.5cm 0cm},clip]{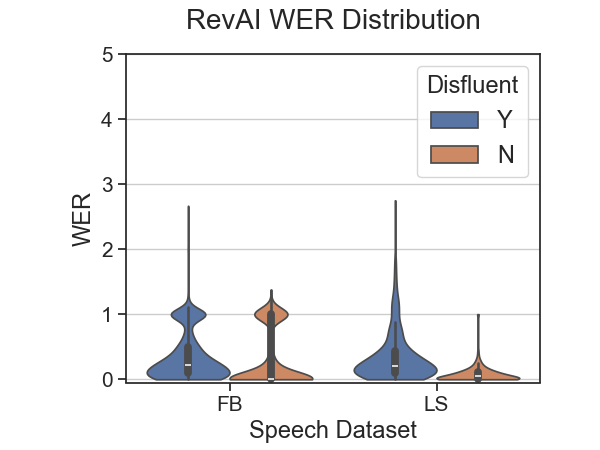}\par
\medskip
\includegraphics[width=.31\textwidth,trim={1.5cm 0cm 1.5cm 0cm},clip]{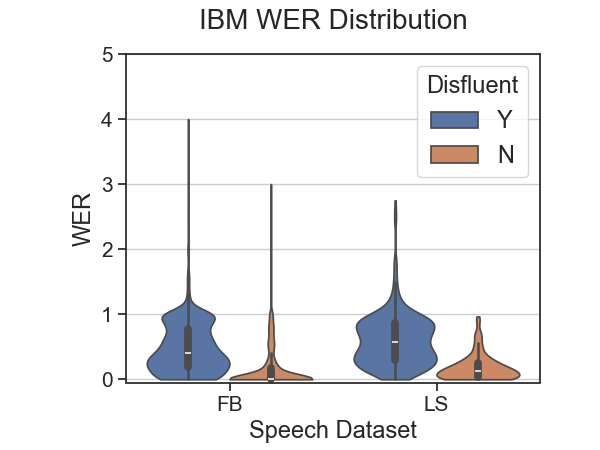}\quad
\includegraphics[width=.31\textwidth,trim={1.5cm 0cm 1.5cm 0cm},clip]{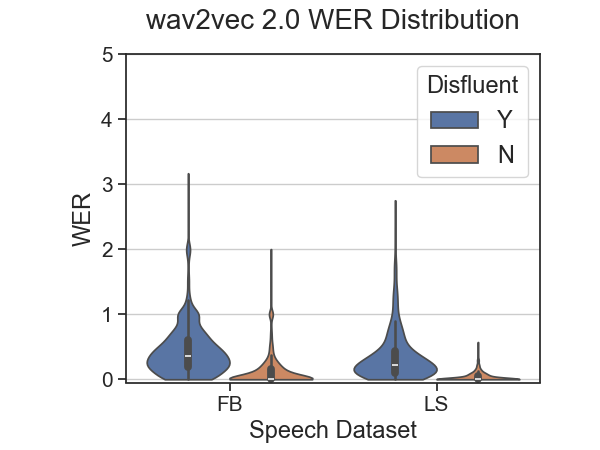}\quad
\includegraphics[width=.31\textwidth,trim={1.5cm 0cm 1.5cm 0cm},clip]{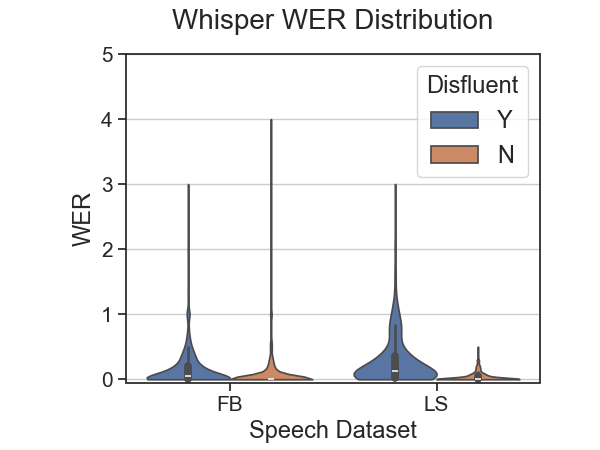}
\caption{Violin plots illustrating the range and distribution of WER for each ASR model on FluencyBank (FB) and LibriSpeech (LS), both with disfluencies (Y) and without disfluencies (N).}
\label{fig:appendix:wers}
\end{figure*}

\begin{figure*}[htp]
\centering
\includegraphics[width=.31\textwidth,trim={1.5cm 0cm 1.5cm 0cm},clip]{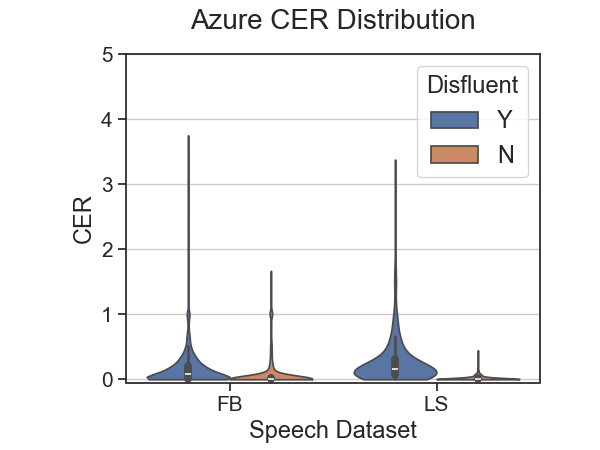}\quad
\includegraphics[width=.31\textwidth,trim={1.5cm 0cm 1.5cm 0cm},clip]{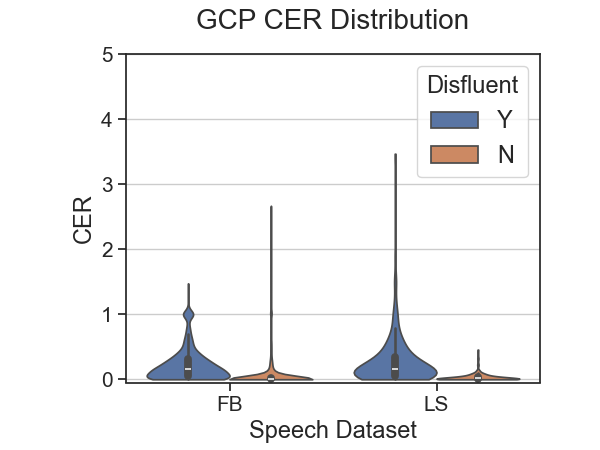}\quad
\includegraphics[width=.31\textwidth,trim={1.5cm 0cm 1.5cm 0cm},clip]{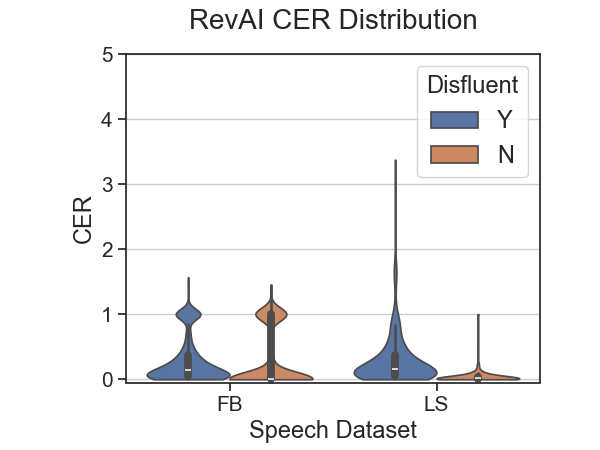}\par
\medskip
\includegraphics[width=.31\textwidth,trim={1.5cm 0cm 1.5cm 0cm},clip]{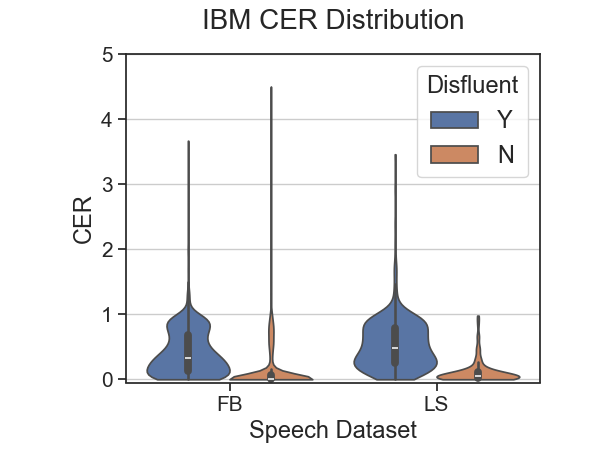}\quad
\includegraphics[width=.31\textwidth,trim={1.5cm 0cm 1.5cm 0cm},clip]{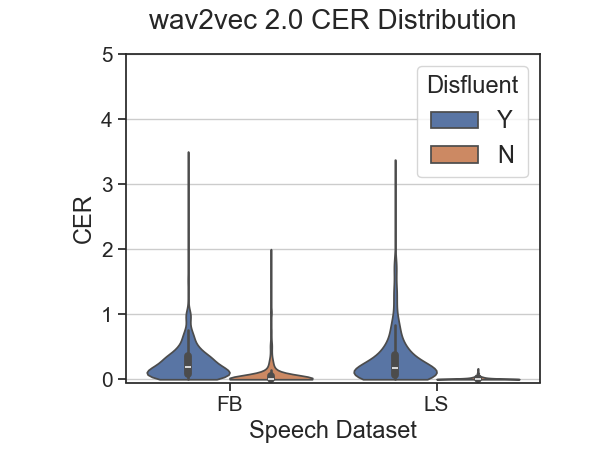}\quad
\includegraphics[width=.31\textwidth,trim={1.5cm 0cm 1.5cm 0cm},clip]{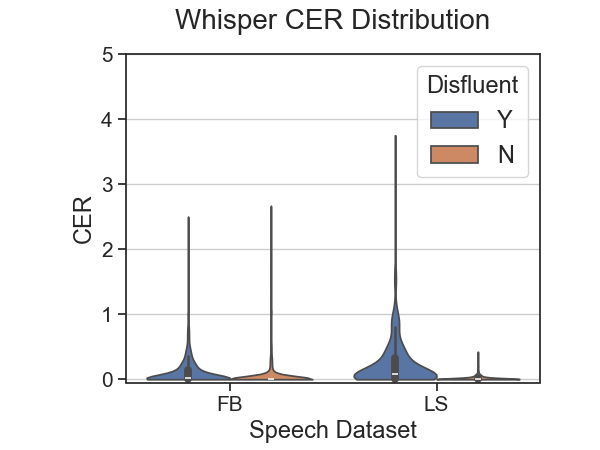}
\caption{Violin plots illustrating the range and distribution of CER for each ASR model on FluencyBank (FB) and LibriSpeech (LS), both with disfluencies (Y) and without disfluencies (N).}
\label{fig:appendix:cers}
\end{figure*}

\begin{figure*}[t]
\centering
\includegraphics[width=.31\textwidth,trim={0cm 0cm 1.5cm 0cm},clip]{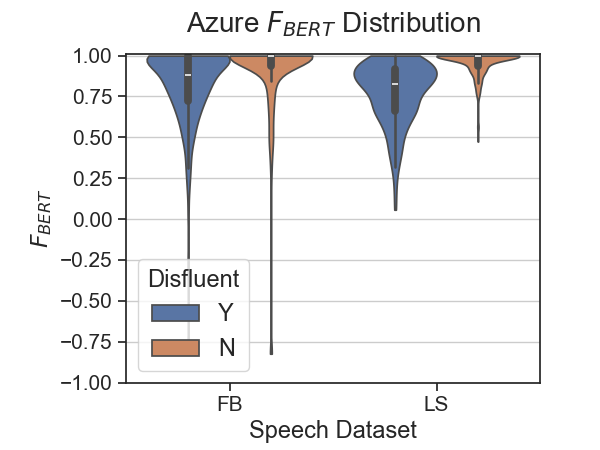}\quad
\includegraphics[width=.31\textwidth,trim={0cm 0cm 1.5cm 0cm},clip]{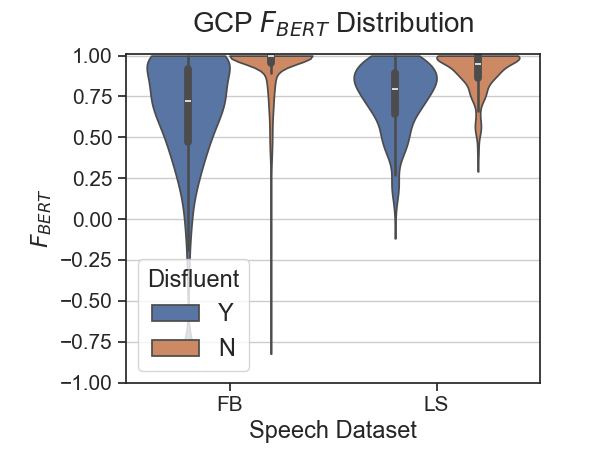}\quad
\includegraphics[width=.31\textwidth,trim={0cm 0cm 1.5cm 0cm},clip]{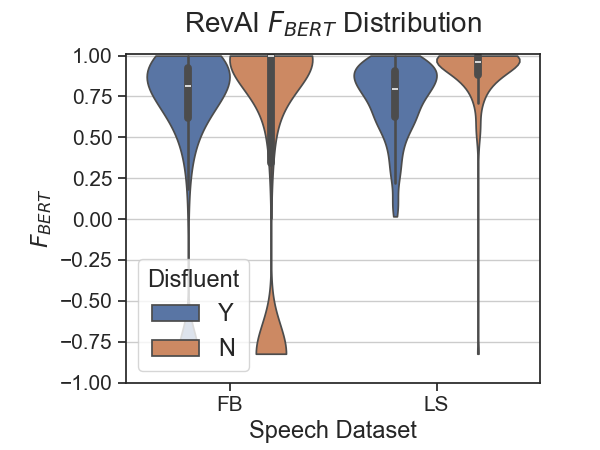}\par
\medskip
\includegraphics[width=.31\textwidth,trim={0cm 0cm 1.5cm 0cm},clip]{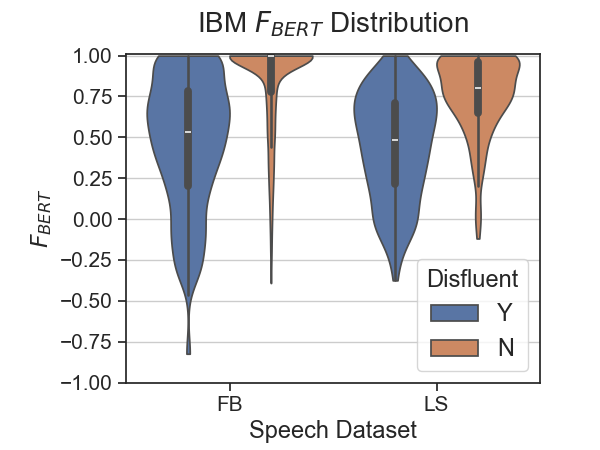}\quad
\includegraphics[width=.31\textwidth,trim={0cm 0cm 1.5cm 0cm},clip]{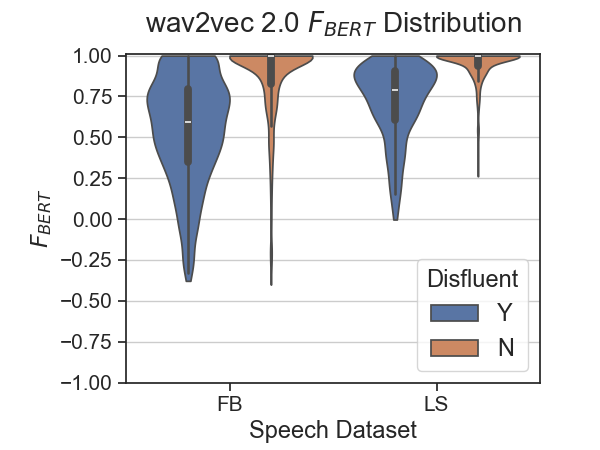}\quad
\includegraphics[width=.31\textwidth,trim={0cm 0cm 1.5cm 0cm},clip]{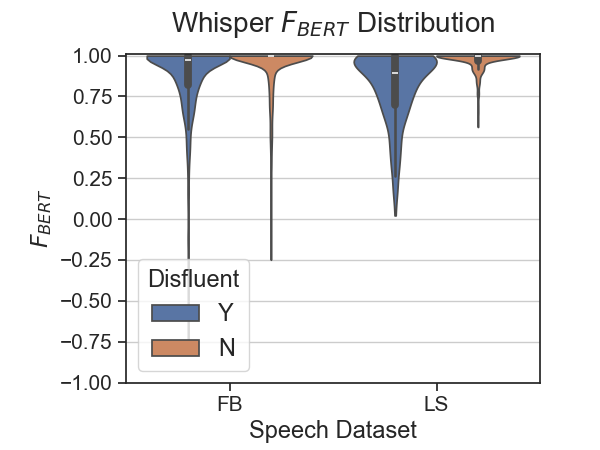}
\caption{Violin plots illustrating the range and distribution of $F_{BERT}$ for each ASR model on FluencyBank (FB) and LibriSpeech (LS), both with disfluencies (Y) and without disfluencies (N).}
\label{fig:appendix:fberts}
\end{figure*}

The distribution of WER, CER, and $F_{BERT}$ scores for all ASR models on real and stuttered speech datasets are presented in Figure \ref{fig:appendix:wers}, Figure \ref{fig:appendix:cers}, and Figure \ref{fig:appendix:fberts}, respectively. We observe that the mean is lower for each ASR when disfluencies are present, specifically in LS-Y and FB-Y, in comparison to LS-N and FB-N without perceived stuttering.

\subsection{Correlation Coefficient Test Results}\label{appendix:spearman-disfluencies}
We test the correlation between degraded ASR accuracy and the presence of disfluent speech using the Spearman rank correlation coefficient, as presented in Table \ref{tbl:pvalue}. Similarly, we also compute the individual correlation coefficients and associated $p$-values for each disfluency type in our synthetic dataset (LS-Y). These results are presented in Table \ref{tbl:stutterevents-detailed}. We find all correlation coefficients, excluding those for interword blocks, statistically significant, with $p < 0.05$. In the case of interword blocks, we find a weak correlation, $r_{WER} < 0.20$, in the GCP, RevAI, and Whisper models without statistical significance, whereas other ASR models have a strong correlation with statistical significance. This is due to the difference in WER being quite small when an interword block is present, as observed in Table \ref{tbl:stutterevents}, with only a 2\% change for RevAI, and 3\% change for GCP and Whisper. This is also seen for the Whisper ASR on the intraword block, where a low correlation $r_{WER} = 0.10$ is observed, as well as only a small decrease, of 2\%, in WER. In comparison, other disfluency types such as word repetitions, interjections, and phrase repetitions show a high correlation with their presence in speech and a decrease in WER. When combining predictions across ASR models, we also observe a statistically significant correlation between all disfluency types, with phrase repetitions having the strongest, and interword blocks having the weakest. 

\end{document}